\documentclass[letterpaper, 10 pt, conference]{ieeeconf}  % Comment this line out if you need a4paper

\IEEEoverridecommandlockouts                              % This command is only needed if 
\usepackage[utf8]{inputenc} % allow utf-8 input
\usepackage[T1]{fontenc}    % use 8-bit T1 fonts
\usepackage{url}            % simple URL typesetting
\usepackage{booktabs}       % professional-quality tables
\usepackage{amsfonts}       % blackboard math symbols
\usepackage{nicefrac}       % compact symbols for 1/2, etc.
\usepackage{microtype}      % microtypography
\usepackage{xcolor}         % colors

\usepackage{graphicx}
\usepackage{xurl}
\usepackage{amssymb}
\usepackage{soul}
\usepackage{amsmath}
\usepackage{mathtools}
\usepackage{amsfonts}
\usepackage{bm}
\usepackage{bbm}
\usepackage{multirow}
\usepackage{multicol}
\usepackage{adjustbox}
\usepackage{threeparttable}
\usepackage{bbding}
\usepackage{colortbl}  
\usepackage{array}  
\usepackage{eso-pic}
\usepackage{wrapfig}
\usepackage{listings}
\usepackage{lipsum}
\usepackage{mdframed}
\usepackage[most]{tcolorbox}
\usepackage{epigraph}
\usepackage{ulem}
\usepackage{longtable}
\usepackage{makecell}
\usepackage{arydshln}
\usepackage{titletoc}

\definecolor{gray}{rgb}{0.5,0.5,0.5}
\definecolor{light-gray}{gray}{0.95}
\definecolor{DarkBlue}{RGB}{72, 116, 203}
\definecolor{royalblue}{rgb}{0.25, 0.41, 0.88}
\definecolor{mediumgray}{rgb}{0.5,0.5,0.5}
\definecolor{AV-red}{HTML}{FF0000}
\definecolor{CBV-purple}{HTML}{9500ff}
\definecolor{BV-blue}{HTML}{001eff}
\usepackage{algpseudocode}
\usepackage{algorithm}
\usepackage{tabularx}
\usepackage{amsmath}
\usepackage{tabularray}
\makeatletter
\newcommand{\multiline}[1]{%
  \begin{tabularx}{\dimexpr\linewidth-\ALG@thistlm}[t]{@{}X@{}}
    #1
  \end{tabularx}
}
\makeatother

\newcommand{\tablestyle}[2]{\setlength{\tabcolsep}{#1}\renewcommand{\arraystretch}{#2}\centering\footnotesize}
\newcommand{\method}{\textit{ForSim}}
\newcommand{\pmsd}[1]{{\color{mediumgray}{\scriptsize $\pm$ #1}}}
\newmdenv[linecolor=light-gray,backgroundcolor=light-gray]{graybox}
\newcommand{\highlighttransp}[2]{%
  \tcbox[colback=#1!50!white, colframe=white,
    enhanced, on line, boxrule=0pt, boxsep=0pt,
    left=1pt, right=1pt, top=1pt, bottom=1pt,
    opacityback=0.3]{#2}%
}

\makeatletter
\let\NAT@parse\undefined
\makeatother
\usepackage{cite}
\usepackage[pagebackref,breaklinks,colorlinks]{hyperref}
\hypersetup{
  linkcolor=DarkBlue,
  citecolor=DarkBlue,
  urlcolor=DarkBlue,
}
\usepackage[capitalize]{cleveref}
\crefname{section}{Sec.}{Secs.}
\Crefname{section}{Section}{Sections}
\Crefname{table}{Table}{Tables}
\crefname{table}{Tab.}{Tabs.}
\crefname{algorithm}{Alg.}{Algs.}
\Crefname{algorithm}{Algorithm}{Algorithms}

\title{\LARGE \bf
ForSim: Stepwise Forward Simulation for Traffic Policy Fine-Tuning
}

\author{Keyu Chen$^{1}$, Wenchao Sun$^{1}$, Hao Cheng$^{1}$, Zheng Fu$^{1}$, Sifa Zheng$^{1}$% <-this % stops a space
\thanks{%
*This work was supported by the National Natural Science Foundation of China under Grant 52572497 and Grant 52221005.
}%
\thanks{%
$^{1}$Keyu Chen, Wenchao Sun, Hao Cheng, Zheng Fu, Sifa Zheng are with the School of Vehicle and Mobility, Tsinghua University, Beijing, China.
}%
}

\begin{document}

\maketitle
\thispagestyle{empty}
\pagestyle{empty}

%%%%%%%%%%%%%%%%%%%%%%%%%%%%%%%%%%%%%%%%%%%%%%%%%%%%%%%%%%%%%%%%%%%%%%%%%%%%%%%%
\begin{abstract}

As the foundation of closed-loop training and evaluation in autonomous driving, traffic simulation still faces two fundamental challenges: covariate shift introduced by open-loop imitation learning and limited capacity to reflect the multimodal behaviors observed in real-world traffic. Although recent frameworks such as RIFT have partially addressed these issues through group-relative optimization, their forward simulation procedures remain largely non-reactive, leading to unrealistic agent interactions within the virtual domain and ultimately limiting simulation fidelity. To address these issues, we propose \method, a stepwise closed-loop forward simulation paradigm. At each virtual timestep, the traffic agent propagates the virtual candidate trajectory that best spatiotemporally matches the reference trajectory through physically grounded motion dynamics, thereby preserving multimodal behavioral diversity while ensuring intra-modality consistency. Other agents are updated with stepwise predictions, yielding coherent and interaction-aware evolution. When incorporated into the RIFT traffic simulation framework, \method\ operates in conjunction with group-relative optimization to fine-tune traffic policy. Extensive experiments confirm that this integration consistently improves safety while maintaining efficiency, realism, and comfort. These results underscore the importance of modeling closed-loop multimodal interactions within forward simulation and enhance the fidelity and reliability of traffic simulation for autonomous driving. Project Page: \href{https://currychen77.github.io/ForSim/}{https://currychen77.github.io/ForSim/}

\end{abstract}

\section{Introduction}
\label{sec:intro}

Traffic simulation forms the foundation of modern autonomous driving, providing the basis for closed-loop training and evaluation~\cite{chen2024review}. It requires simulating multi-agent behavior over time in a way that reflects real-world driving behaviors within a closed-loop environment. This introduces two central challenges: \textit{covariate shift} and \textit{multimodality}.

Covariate shift is a long-standing issue in open-loop imitation learning (IL), where models trained on offline expert demonstrations often struggle when deployed in closed-loop settings due to compounding errors and distributional drift. A classical solution is online learning~\cite{ross2011dagger}, which unrolls the policy and queries an expert to generate new demonstrations~\cite{peng2023PVP}. However, in multi-agent traffic simulation, frequent expert querying becomes prohibitively expensive and unscalable. To address this challenge, recent studies integrate imitation learning (IL) with reinforcement learning (RL) under a closed-loop paradigm~\cite{lu2023imitationisnotenough, zhang2023rtr, peng2024improving}. This hybrid approach leverages the complementary strengths of the two paradigms: IL provides sample-efficient initialization by distilling expert demonstrations, while RL enables policies to adapt through interaction-driven feedback in closed-loop execution. By jointly exploiting these advantages, such methods reduce covariate shift by grounding policy updates in causal links between observations, actions, and rewards, while leveraging expert data to accelerate the learning of realistic driving behaviors. In contrast to IL-RL hybrid approaches, CAT-K~\cite{zhang2025catk} adopts a purely supervised strategy through closed-loop fine-tuning, aligning each rollout with the closest among top-k expert trajectories. This alignment preserves the rollout's consistency with ground-truth (GT) demonstrations, thereby maintaining valid supervision signals throughout the closed-loop process. However, despite this improvement, such methods typically rely on optimizing the policy using only the most likely rollout stored in the buffer, which corresponds to a single behavior modality over the rollout horizon. Consequently, they struggle to capture the multimodal driving behaviors in real-world traffic. Addressing this limitation requires not only a dataset with coverage of diverse driving modalities but also a policy capable of maintaining exploration during closed-loop execution to avoid mode collapse. This places substantial demands on both data diversity and the multi-modal modeling capacity of the policy.

\begin{figure}[t]
  \centering
  \includegraphics[width=0.49\textwidth]{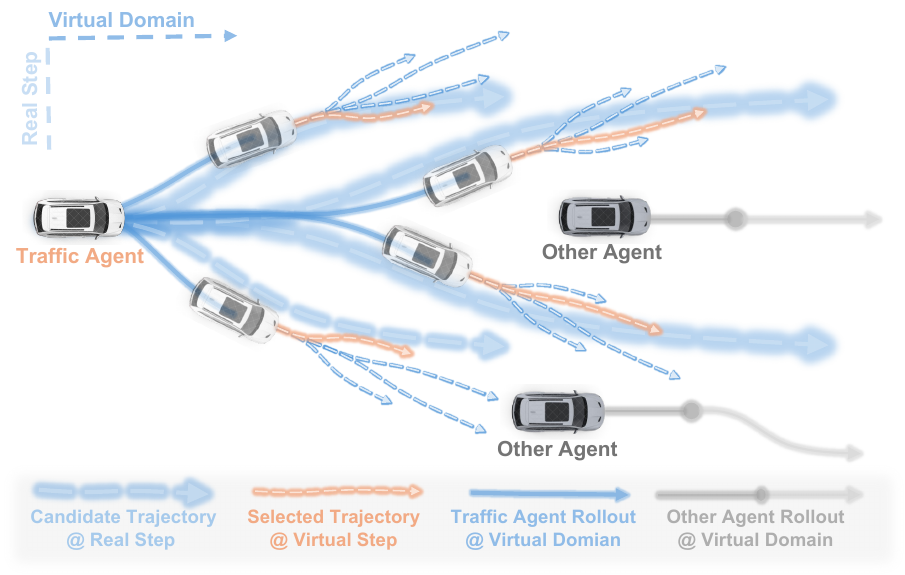}
  \vspace{-5mm}
  \caption{\textbf{\method} introduces stepwise unrolling of multimodal candidate trajectories under closed-loop dynamics. At each virtual timestep, the traffic agent selects the spatiotemporal aligned trajectory to preserve modality consistency, while propagation through the PID controller and kinematic bicycle model ensures physical plausibility. Other agents follow stepwise predictions, ensuring interactive and coherent evolution.}
  \label{fig:intro}
  \vspace{-5mm}
\end{figure}

To leverage the strengths of both IL and RL, RIFT~\cite{chen2025rift} first performs open-loop IL pre-training on real-world datasets to capture realistic driving behaviors, and then fine-tunes the policy via group-relative optimization to mitigate covariate shift while preserving multimodality. However, RIFT conducts non-reactive forward simulations: only the first step is closed-loop, while subsequent steps are rolled out in open-loop across all candidate trajectories. This leads to unrealistic agent interactions in the virtual time domain, limiting the fidelity of the forward simulation.

Building on these observations, we introduce \method, a stepwise closed-loop forward simulation paradigm designed to capture the interaction within the virtual forward simulation. In the real-time domain, each traffic agent generates multiple candidate trajectories through the traffic policy, representing distinct plausible modalities. \method\ performs forward simulation independently for each candidate, treating it as a reference trajectory in an associated virtual domain. At every virtual timestep, the agent propagates the trajectory that best spatiotemporally aligns with this reference, ensuring intra-modality consistency throughout the rollout. Propagation adheres to physically grounded dynamics, ensuring physical plausibility and dynamic coherence. Other agents are propagated along predicted trajectories that are updated at each virtual timestep, ensuring responsiveness to evolving interactions. This iterative procedure guarantees that all candidate modalities evolve coherently under closed-loop conditions, simultaneously preserving multimodal behavioral diversity, intra-modality consistency, and physical plausibility. The resulting multimodal rollouts are evaluated in a stepwise fashion within the RIFT framework, enabling fine-grained optimization of traffic policy through group-relative objectives.

Our contributions are summarized as follows:

\begin{itemize}
    \item We propose \method, a stepwise closed-loop forward simulation paradigm that ensures behavioral multimodality, intra-modality consistency, and physical plausibility, thereby enabling fine-grained optimization of traffic policy within the RIFT framework.
    \item We demonstrate that \method\ outperforms existing forward simulation paradigms by modeling closed-loop multimodal interactions, while further enhancing the realism and reliability of traffic simulation.
\end{itemize}
\section{Related Work}
\label{sec:relatedwork}

\textbf{Traffic Simulation.} Recent advances in traffic simulation have explored diverse generative architectures—including VAEs~\cite{trafficsim, xu2023bits}, diffusion models~\cite{chitta2024sledge, zhou2024Nexus}, and next-token prediction~\cite{wu2024smart, philiontrajeglish}—to improve realism~\cite{tan2021scenegen} and long-horizon robustness~\cite{yang_infgen, peng2025infgen}. At the same time, controllability has been pursued via cost-based conditioning~\cite{zhong2023CTG}, language prompts~\cite{zhong2023CTG++, tan2024prosim}, guided sampling~\cite{lu2024scenecontrol}, and retrieval-based generation~\cite{ding2024realgen}, enabling user-aligned traffic scenario generation~\cite{lin2024ccdiff,chen2025frea}. Yet, most approaches suffer from the covariate shift problem caused by the mismatch between open-loop training and closed-loop deployment, which further undermines long-term stability. Closed-loop fine-tuning strategies provide partial remedies, yet each comes with trade-offs. Hybrid IL/RL methods~\cite{zhang2023rtr, peng2024improving, lu2023imitationisnotenough} enhance robustness but often compromise realism due to reward design challenges. Supervised approaches such as CAT-K~\cite{zhang2025catk} achieve strong performance but depend on scarce expert demonstrations. RLHF~\cite{cao2024trafficrlhf} improves human alignment but requires costly feedback and suffers from reward model instability. RIFT~\cite{chen2025rift} mitigates covariate shift while preserving multimodality via group-relative optimization, but its reliance on non-reactive forward simulation limits the realism of closed-loop agent interactions within the virtual domain.

\begin{figure}[t]
  \centering
  \includegraphics[width=0.48\textwidth]{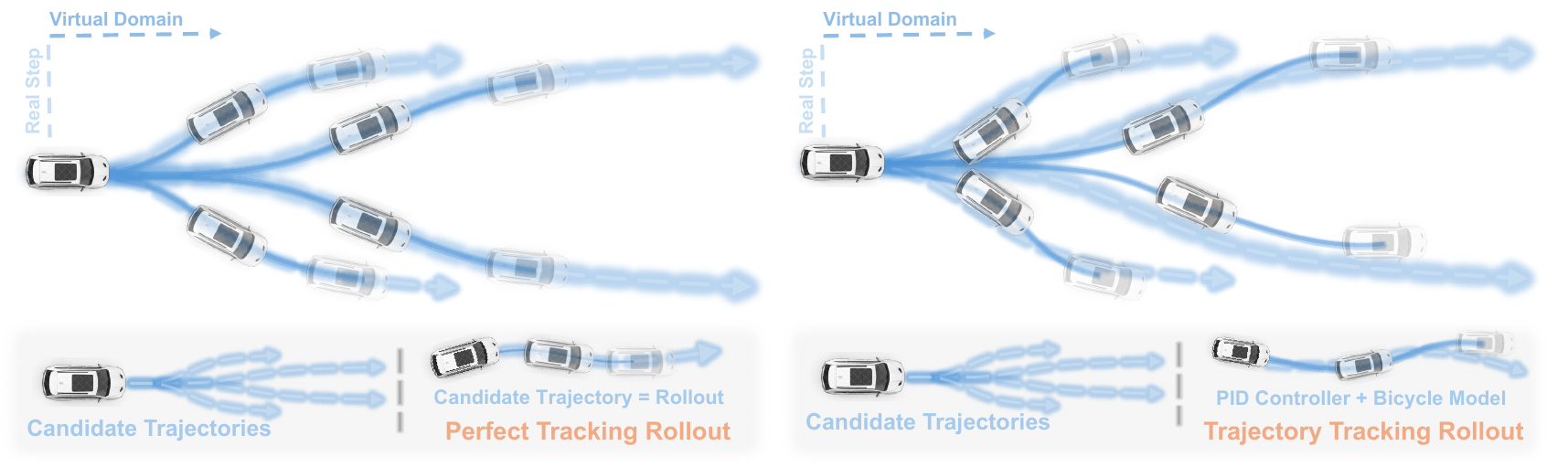}
  \vspace{-6mm}
  \caption{\textbf{Typical rollout paradigms.} The left panel depicts \textit{Perfect Tracking}, in which the vehicle strictly follows the planned trajectory. The right panel depicts \textit{Trajectory Tracking}, which employs a controller and kinematic bicycle model to follow the trajectory under dynamic constraints.}
  \label{fig:basic-rollout}  
  \vspace{-5mm}
\end{figure}

\textbf{Forward Simulation.} In autonomous driving, forward simulation denotes virtually unrolling the future evolution of the center vehicle and other agents conditioned on a given trajectory, serving to evaluate the planned trajectories and bridging the gap between planning and execution. As illustrated in \Cref{fig:basic-rollout}, typical rollout paradigms for the center vehicle include \textit{Perfect Tracking}, where the vehicle strictly follows the planned trajectory, and \textit{Trajectory Tracking}, where a controller and kinematic bicycle model~\cite{kinematicbicyclemodel} are used to follow the trajectory under dynamic constraints. In practice, Trajectory Tracking is the dominant paradigm, and existing approaches primarily diverge in their modeling of responses from other agents. PDM-Closed~\cite{dauner2023tuplan_garage} selects among candidate trajectories by unrolling each one with Trajectory Tracking, while forecasting other agents under a constant-velocity assumption~\cite{constantvelocity}. PlanTF~\cite{cheng2024planTF} and Pluto~\cite{cheng2024pluto} integrate forward simulation into training or post-processing to better align planned trajectories and executed rollouts, but assume other agents follow log-replay. NAVSIM~\cite{dauner2024navsim} achieves pseudo-closed-loop evaluation on the nuPlan dataset~\cite{caesar2021nuplan} by combining trajectory unrolling with log-replay~\cite{dauner2024navsim} or rule-based updates~\cite{cao2025navsimv2}. Despite these advances, most approaches remain essentially open-loop, unrolling fixed trajectories without replanning and propagating other agents under static forecasts, neglecting dynamic inter-agent interactions. This discrepancy from the real-world plan–execute–replan cycle limits the fidelity and reliability of forward simulation.
\section{Methodology}
\label{sec:method}

As outlined in \Cref{sec:intro}, this work aims to mitigate covariate shift in traffic simulation while preserving multimodal diversity and forward simulation reliability. A suitable foundation for this objective is RIFT~\cite{chen2025rift}, which couples IL pre-training, capturing realistic behaviors with route-level controllability, with RL fine-tuning in a high-fidelity simulator to reduce covariate shift and enhance style controllability.

Despite these strengths, RIFT adopts a non-reactive forward simulation paradigm, equivalent to PDM-Closed~\cite{dauner2023tuplan_garage}, where only the first step is closed-loop while subsequent steps remain open-loop. This neglects inter-agent interactions and limits long-term realism. We address this limitation by refining the forward simulation component to enable closed-loop, interaction-aware rolling of candidate trajectories, thereby ensuring more faithful evaluation of multimodal candidate trajectories and ultimately improving the realism, reliability, and controllability of traffic simulations.

\begin{figure*}[t]
  \centering
  \includegraphics[width=1.0\textwidth]{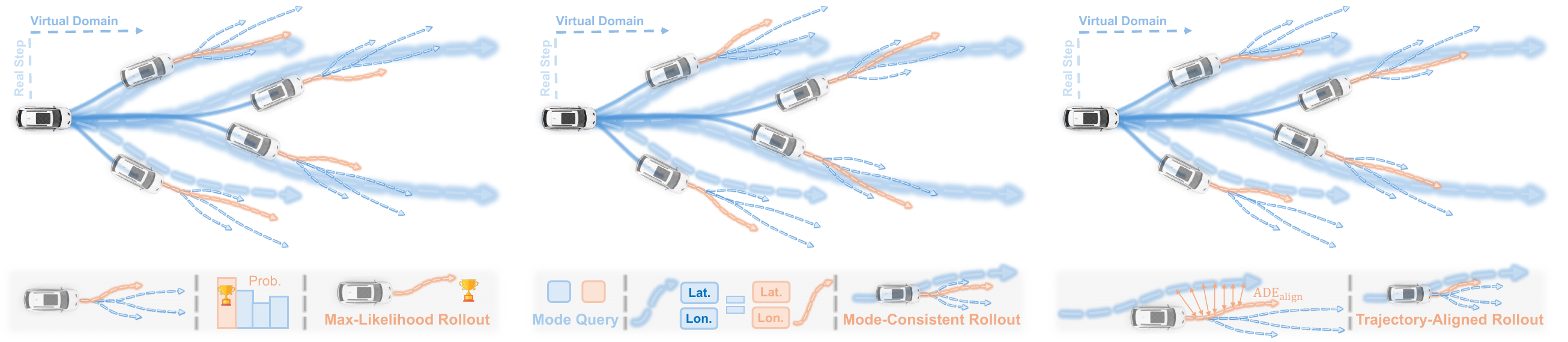}
  \vspace{-5mm}
  \caption{\textbf{Illustration of three stepwise rollout paradigms for traffic agents:} Max-Likelihood Rollout, Mode-Consistent Rollout, and Trajectory-Aligned Rollout. Max-likelihood Rollout rapidly collapses multimodal diversity, while Mode-Consistent Rollout suffers from accumulated misalignment across virtual states. Trajectory-Aligned Rollout, in contrast, selects at each step the candidate trajectory closest to its initial mode—measured by Average Displacement Error (ADE)—thereby preserving multimodal fidelity and yielding physically coherent rollouts.} 
  \label{fig:step-rollout}
  \vspace{-5mm}
\end{figure*}

\subsection{Preliminaries}
\label{subsec:preliminary}
Following RIFT~\cite{chen2025rift}, we construct an AV-centric closed-loop simulation in Carla~\cite{carla}. The critical background vehicles (CBVs)—those most likely to interact with the AV—are controlled by the traffic policy, and we interchangeably refer to each CBV as a traffic agent in the following discussion. For each traffic agent, the policy takes as input its current feature $F_{\mathrm{cbv}}$, historical features of neighboring vehicles $F_{\mathrm{neighbor}}$, and vectorized map features $F_{\mathrm{map}}$. These are encoded into embeddings $E_{\mathrm{cbv}} \in \mathbb{R}^{1 \times D}$, $E_{\mathrm{neighbor}} \in \mathbb{R}^{N_{\mathrm{neighbor}} \times D}$, and $E_{\mathrm{map}} \in \mathbb{R}^{N_{\mathrm{map}} \times D}$, where $N_{\mathrm{neighbor}}$ and $N_{\mathrm{map}}$ denote the number of neighboring vehicles and map elements, respectively, and $D$ is the embedding dimension. The resulting embeddings are aggregated through Transformer encoder blocks to form the CBV-centric scene embedding $E_{\mathrm{enc}}$.

To capture multimodal driving behaviors, the policy further adopts a longitudinal–lateral decoupling mechanism~\cite{cheng2024pluto}. Specifically, high-level lateral queries $Q_{\mathrm{lat}}\in\mathbb{R}^{N_{\mathrm{ref}} \times D}$ are constructed from reference line information, while learnable longitudinal queries $Q_{\mathrm{lon}}\in\mathbb{R}^{N_{\mathrm{lon}} \times D}$ represent diverse longitudinal anchors. These are combined into multimodal navigation queries $Q_{\mathrm{nav}}\in\mathbb{R}^{N_{\mathrm{ref}} \times N_{\mathrm{lon}} \times D}$, which interact with the scene embedding $E_{\mathrm{enc}}$ through Transformer decoder blocks. The decoder's final output $Q_{\mathrm{dec}}$ is passed through MLP heads to generate a set of candidate trajectories $\mathcal{T}\in\mathbb{R}^{N_{\mathrm{ref}} \times N_{\mathrm{lon}} \times T \times 6}$ along with their confidence scores $\mathcal{S}\in\mathbb{R}^{N_{\mathrm{ref}} \times N_{\mathrm{lon}}}$, where each trajectory point $\tau_i^t$ encodes $[p_x, p_y, \cos\theta, \sin\theta, v_x, v_y]$. Among these candidates, the trajectory with the highest confidence score is executed by a PID controller, ensuring that CBVs remain in closed-loop interaction with the environment.

In parallel, all candidate trajectories are unrolled through the forward simulation to generate rollouts $\widetilde{\mathcal{T}} = \{\widetilde{\tau}_i\}_{i=1}^{N_{\mathrm{ref}} \times N_{\mathrm{lon}}}$, which form the foundation for modality evaluation and subsequently fine-tune the traffic policy through group-relative optimization. Since our work focuses on forward simulation, the core challenge is to unroll all candidate trajectories in a way that faithfully captures real-world interactions and transitions, thereby providing a reliable foundation for rollout evaluation and policy optimization.

To address this challenge, we decompose the forward simulation into two components: (i) stepwise virtual rollout of the traffic agent and (ii) stepwise prediction propagation of other agents, which together define how candidate trajectories are unrolled under closed-loop interaction.

\subsection{Traffic Agent Stepwise Virtual Rollout}
\label{subsec:center_rollout}

In contrast to Perfect Tracking, which directly reuses the planned trajectory, and Trajectory Tracking, which rigidly tracks a fixed trajectory without adaptation—both illustrated in \Cref{fig:basic-rollout}—our approach embraces a \textit{stepwise planning–execution–replanning} process that dynamically adapts to real-time state changes. This framework more faithfully captures real-world driving dynamics, where agents continually revise their plans in response to evolving states and interactions. Concretely, at the initial real-world time step ($t=0$) we generate $N_{\mathrm{ref}}\times N_{\mathrm{lon}}$ candidate trajectories and forward-simulate each for one step, yielding $N_{\mathrm{ref}}\times N_{\mathrm{lon}}$ rollout branches at virtual time $\tilde{t}{=}1$. 
Subsequently, at each virtual step $\tilde{t}$, each branch selects a candidate index $(i_{\tilde{t}},j_{\tilde{t}})$ based on a specified paradigm and propagates its state by tracking the selected trajectory via a PID controller coupled with a kinematic bicycle model:
\begin{equation}
\label{eq:virtual_prop}
x_{\tilde{t}+1}
= \mathcal{F}_{\mathrm{PID+Bike}}\!\big(\,x_{\tilde{t}},\; \tau_{(i_{\tilde{t}},j_{\tilde{t}})}(x_{\tilde{t}})\,\big),
\quad \tilde{t}=1,\ldots,T\!,
\end{equation}
where $x_{\tilde{t}}$ denotes the virtual state at $\tilde{t}$ and $\mathcal{F}_{\mathrm{PID+Bike}}$ represents one-step execution via PID tracking followed by kinematic bicycle propagation. 
To instantiate this framework, we explore three stepwise rollout paradigms, illustrated in \Cref{fig:step-rollout}: the \textit{Max-Likelihood Rollout}, the \textit{Mode-Consistent Rollout}, and the \textit{Trajectory-Aligned Rollout}.

\textbf{Max-Likelihood Rollout.} 
At each virtual time step $\tilde{t}$, each branch selects the candidate with the highest confidence score predicted at the current virtual state:
\begin{equation}
(i_{\tilde{t}},j_{\tilde{t}}) \;=\; \arg\max_{i,j}\; s_{(i,j)}(x_{\tilde{t}}).
\end{equation}
This strategy ensures that each branch consistently follows the most likely mode throughout the rollout. Nevertheless, this strategy inevitably causes branches to converge toward the dominant mode. Because the policy is robust to small state perturbations, initial deviations introduced by diverse candidates are often ignored, ultimately collapsing the intended multimodal behavior.

\textbf{Mode-Consistent Rollout.} 
An alternative strategy enforces each branch to remain in the same candidate mode selected at the initial step from $t=0$ to $\tilde{t}=1$. 
Specifically, if a branch is seeded at $\tilde{t}=1$ with candidate index $(i^{\mathrm{ref}},j^{\mathrm{ref}})$, then it maintains $(i_{\tilde{t}},j_{\tilde{t}})=(i^{\mathrm{ref}},j^{\mathrm{ref}})$ at all subsequent virtual steps and propagates accordingly via $\mathcal{F}_{\mathrm{PID+Bike}}$ accordingly. 
While this approach guarantees explicit mode consistency, it introduces misalignment: trajectories corresponding to the same candidate mode at different virtual states are not perfectly coherent, and the resulting accumulation of deviations can cause unintended mode shifts—ultimately compromising rollout stability.

\textbf{Trajectory-Aligned Rollout.} 
To address the limitations of the previous strategies, we propose the Trajectory-Aligned paradigm. Each branch is still initialized at virtual time $\tilde{t}{=}1$ using a specific candidate index $(i^{\mathrm{ref}}, j^{\mathrm{ref}})$, corresponding to the trajectory selected during the initial transition from $t{=}0$ to $\tilde{t}{=}1$. This reference trajectory remains fixed throughout the rollout. At every subsequent virtual step $\tilde{t} \geq 2$, the branch dynamically selects the candidate whose trajectory—after temporal alignment—is closest to the fixed reference, as measured by the Average Displacement Error (ADE):
\begin{equation}
\label{eq:closest_select}
(i_{\tilde{t}},j_{\tilde{t}}) 
\;=\;
\arg\min_{(i,j)}\;
\mathrm{ADE}_{\mathrm{align}}\!\Big(
\tau_{(i,j)}(x_{\tilde{t}}),\;
\tau_{(i^{\mathrm{ref}}, j^{\mathrm{ref}})}
\Big),
\end{equation}
where $\mathrm{ADE}_{\mathrm{align}}$ denotes ADE after temporal alignment of the reference trajectory to the virtual evaluation window.

This formulation enforces fidelity to the original modality while avoiding mode collapse toward the Max-Likelihood Rollout and the temporal drift inherent in Mode-Consistent Rollout. Coupled with state propagation through $\mathcal{F}_{\mathrm{PID+Bike}}$, Trajectory-Aligned Rollout preserves multimodal consistency and guarantees rollouts that are both physically plausible and dynamically coherent, making it the most effective paradigm.

\begin{figure}[t]
  \centering
  \includegraphics[width=0.48\textwidth]{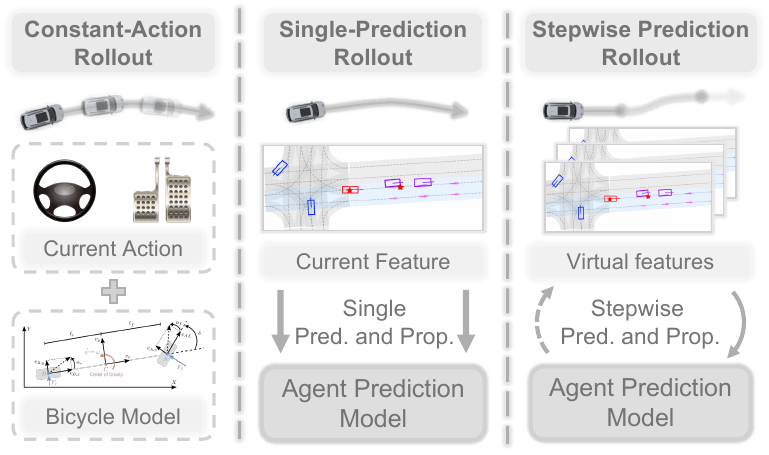}
  \vspace{-5mm}
  \caption{\textbf{Other agents rollout paradigms:} Constant-Action Rollout propagates the current action in an open-loop manner; Single-Prediction Rollout yields more plausible trajectories but remains open-loop; and Stepwise Prediction Rollout enables closed-loop, interaction-aware rollout.}
  \label{fig:agent-pred}
  \vspace{-5mm}
\end{figure}

\subsection{Other Agents Stepwise Prediction Rollout}
\label{subsec:agents_rollout}

In addition to unrolling the traffic agent, accurately simulating other agents is critical for realistic interactions. We explore three paradigms, illustrated in \Cref{fig:agent-pred}.

\textbf{Constant-Action Rollout.} The most commonly adopted baseline assumes that each other agent maintains its current action throughout the entire rollout horizon. The recorded control input at the current time step is propagated forward using a kinematic bicycle model. While this method ensures physical feasibility, it operates in a purely open-loop fashion and fails to capture interaction-induced behavior shifts, often resulting in unrealistic agent responses.

\textbf{Single-Prediction Rollout.} A more advanced alternative utilizes a predictor to forecast agent trajectories based on the current states. This predicted trajectory is then directly propagated through time. Compared to Constant-Action Rollout, this method yields more plausible estimates of future states. Since predictions are generated only at initialization, the simulation remains open-loop and fails to capture interactive behaviors in response to evolving traffic dynamics.

\textbf{Stepwise Prediction Rollout.} To address these limitations, we employ a Stepwise Prediction paradigm in which other agents update their predicted trajectories at every virtual time step, conditioning on the evolving virtual state. Each updated prediction is then propagated forward one step. This design synchronizes the planning frequency of other agents with that of the traffic agent, enabling responsive behaviors and interaction-aware updates. Consequently, it facilitates closed-loop dynamics and significantly improves the realism of inter-agent interactions over long horizons.

At the core of this paradigm lies an agent prediction model that produces horizon-$T$ trajectories for all other agents, denoted as $\hat{\mathbf{y}} \in \mathbb{R}^{N\times T\times 6}$, where each state encodes the agent’s position, orientation, and velocity. Ground-truth supervision derives from future observations over the first $T_f$ steps, forming the target trajectories $\mathbf{y}^\star$. To balance fidelity to ground-truth data with physical realism, we integrate multiple complementary loss terms into a composite objective:
\begin{equation}
\label{eq:pred_loss}
\mathcal{L}_{\text{pred}}
= w_{\text{anchor}} \mathcal{L}_{\text{anchor}}
+ w_{\text{kin}} \mathcal{L}_{\text{kin}}
+ w_{\text{smooth}} \mathcal{L}_{\text{smooth}},
\end{equation}
where the individual loss terms are defined as
\begin{align}
\mathcal{L}_{\text{anchor}} &= 
\operatorname{SmoothL1}\!\left(\hat{\mathbf{y}}_{0:T_f-1}, \mathbf{y}^\star\right), \\
\mathcal{L}_{\text{kin}} &= 
\operatorname{SmoothL1}\!\left(\mathbf{v}_t,\; 
\frac{\mathbf{p}_{t+1}-\mathbf{p}_{t-1}}{2\Delta t}\right), \\
\mathcal{L}_{\text{smooth}} &= 
\|\Delta \mathbf{v}_t\|_1 + \|\Delta^2 \mathbf{v}_t\|_1,
\end{align}
with $\mathbf{p}_t=(x_t,y_t)$, $\mathbf{v}_t=(v^x_t,v^y_t)$, 
$\Delta \mathbf{v}_t = \mathbf{v}_{t+1}-\mathbf{v}_t$, and 
$\Delta^2 \mathbf{v}_t=\Delta \mathbf{v}_{t+1}-\Delta \mathbf{v}_t$.
This objective balances short-term accuracy, kinematic consistency, and temporal smoothness, resulting in predictions that are both physically plausible and stable for long-horizon closed-loop simulations. We set $w_{\text{anchor}}=1.0$, $w_{\text{kin}}=0.5$, and $w_{\text{smooth}}=0.5$.

\subsection{Group-Relative Policy Fine-Tuning}

Building on the results of forward simulation, we fine-tune the traffic policy in a closed-loop manner by leveraging the group-relative optimization principle introduced in RIFT~\cite{chen2025rift}. For each traffic agent's real-world state $s$, $G=N_{\mathrm{ref}} \times N_{\mathrm{lon}}$ candidate trajectories are forward-simulated to generate rollout branches, which are subsequently evaluated using a stepwise reward model to obtain rollout-level returns $\mathcal{R} = \{r_i\}_{i=1}^G$. Formally, given a set of rollouts $\widetilde{\mathcal{T}} = \{\widetilde{\tau}_i\}_{i=1}^G$, we define:
\begingroup
\small
\begin{align}
\label{eq:group_reward_advantage} 
r_i = \sum_{t=0}^T \gamma^t \left[\mathrm{RM}(\widetilde{\tau}_i^t)\right],
\hat{A}_i = \frac{r_i - \mathrm{mean}(\mathcal{R})}{\mathrm{std}(\mathcal{R})}, \forall i = 1, \dots, G
\end{align}
\endgroup
where $\hat{A}_i$ represents its normalized group-relative advantage. This normalization promotes high-return behaviors while preserving the multimodality inherent in the candidate set.

\begin{algorithm}[t!]
\small
\caption{Closed-Loop Traffic Policy Optimization}
\label{alg:policy_training}
\begin{algorithmic}[1]
    \State \textbf{Input}: IL pre-trained traffic policy $\pi_{\theta_{\text{init}}}$, IL pre-trained predictor $\mathcal{M}_{\text{pred}}$, buffer $\mathcal{D}$
    \State Policy $\pi_\theta \leftarrow \pi_{\theta_{\text{init}}}$
    \For{$\text{iteraction}=1,\dots,I$} \Comment{RL fine-tuning}
        \State Update the old policy $\pi_{\theta_{old}} \leftarrow \pi_{\theta}$
        \While{$\mathcal{D}$ not full} \Comment{Collect rollout data}
            \For{$\text{step}=1,\dots,T$}
                \State \multiline{Generate $\{\tau_i\}_{i=1}^{G}$ from $\pi_{\theta_{old}}$ \Comment{Policy inference}}
                \State \multiline{Derive $\{\widetilde{\tau}_i\}_{i=1}^{G}$ via \cref{alg:for_sim} \Comment{Forward simulation}}
                \State \multiline{Compute $\{r_i\}_{i=1}^{G}$, $\{\hat{A}_i\}_{i=1}^{G}$ for each $\widetilde{\tau}_i$ with \cref{eq:group_reward_advantage}} 
                \State \multiline{Store transition into buffer $\mathcal{D}$}
            \EndFor
        \EndWhile
        \For{\method\ $\text{iteration}=1,\dots,\mu$} \Comment{Fine-tuning}
            \State \multiline{Sample mini-batch transitions from the buffer $\mathcal{D}$}
            \State \multiline{Update predictor $\mathcal{M}_{\text{pred}}$ via \cref{eq:pred_loss} \Comment{Train predictor}}
            \State \multiline{Update traffic policy $\pi_\theta$ via \cref{eq:rift_loss} \Comment{Train policy}}
        \EndFor
    \EndFor
    \State \textbf{Output}: RL fine-tuned traffic policy
\end{algorithmic}
\end{algorithm}

\begin{algorithm}[t!]
\small
\caption{\method: Stepwise Forward Simulation}
\label{alg:for_sim}
\begin{algorithmic}[1]
    \State \textbf{Input}: Reference trajectories $\{\tau_i\}_{i=1}^{G}$, traffic policy $\pi_{\theta_{\text{init}}}$, predictor $\mathcal{M}_{\text{pred}}$
    \For{$\tau_i \in \{\tau_i\}_{i=1}^{G}$}
        \State \multiline{Propagate $\tau_i$ via $\mathcal{F}_{\mathrm{PID+Bike}}$ to obtain $x_1^{\text{center}}$}
        \State \multiline{Predict other agents via $\mathcal{M}_{\text{pred}}$ to obtain $x_1^{\text{agents}}$}
        \State \multiline{Record initial rollout state $x_1 = (x_1^{\text{center}}, x_1^{\text{agents}})$}
        \For{$\text{Virtual step}\ \tilde{t}=1,\dots,T$} \Comment{Stepwise rollout}
            \State \multiline{Generate trajectories from $x_{\tilde{t}}$, select via \cref{eq:closest_select}, then propagate via \cref{eq:virtual_prop} to obtain $x_{\tilde{t}+1}^{\text{center}}$ \Comment{traffic agent}}
            \State \multiline{Predict $x_{\tilde{t}+1}^{\text{agents}}$ via $\mathcal{M}_{\text{pred}}(x_{\tilde{t}})$ \Comment{other agents}}
            \State \multiline{Record next rollout state $x_{\tilde{t}+1} = (x_{\tilde{t}+1}^{\text{center}}, x_{\tilde{t}+1}^{\text{agents}})$}
        \EndFor
    \EndFor
    \State \textbf{Output}: Rollouts $\{\widetilde{\tau}_i\}_{i=1}^{G}$
\end{algorithmic}
\end{algorithm}

To update the policy, we adopt the group-relative optimization framework of RIFT, which employs a dual-clip mechanism to bound policy updates, stabilize training, and preserve alignment with user-preferred styles. The resulting optimization objective is:

\begingroup
\scriptsize
\begin{equation}
\begin{aligned}
\label{eq:rift_loss}
\mathcal{J}(\theta)
&=\mathbb{E}_{s\sim\mathcal D}\!\left[
\frac{1}{G}\sum_{i=1}^{G}
\;\psi\big(\rho_i(\theta),\,\hat A_i\big)
\right], \quad
\rho_i(\theta) = \frac{\pi_\theta(\tau_i | s)}{\pi_{\theta_{\mathrm{old}}}(\tau_i | s)}, \\
\psi(\rho,\hat A) &=
\begin{cases}
\min\!\big(\rho\,\hat A,\ \mathrm{clip}(\rho,1-\epsilon,1+\epsilon)\,\hat A\big), & \hat A\ge 0,\\
\max\!\Big(\min\!\big(\rho\,\hat A,\ \mathrm{clip}(\rho,1-\epsilon,1+\epsilon)\,\hat A\big),\ c\,\hat A\Big), & \hat A<0
\end{cases},
\end{aligned}
\end{equation}
\endgroup
where $\rho_i(\theta)$ denotes the trajectory-level likelihood ratio between the updated and old traffic policies.

In this manner, the traffic policy is fine-tuned directly on forward-simulated rollouts, enabling improved controllability, mitigating covariate shift, and enhancing training stability in a closed-loop setting. The overall training pipeline (Algorithm~\ref{alg:policy_training}) repeatedly invokes the forward simulation module (Algorithm~\ref{alg:for_sim}) to generate virtual rollouts for policy updates.

\section{Experiment}
\label{sec:exp}

In this section, we empirically evaluate the effectiveness of \method\ in enhancing the realism and controllability of traffic simulations. Our experiments are designed to answer the following key research questions: \textbf{Q1}: Does the closed-loop forward simulation in \method\ improve traffic simulation quality compared to baseline approaches? \textbf{Q2}: What are the individual contributions of the rollout paradigms within \method\ to the overall performance? \textbf{Q3}: Why do the proposed rollout paradigms work effectively in practice?

We begin by outlining the experimental setup, including the simulation platform, evaluation metrics, and baselines. We then conduct a comprehensive empirical comparison, demonstrating the superiority of \method\ under various quantitative and qualitative criteria. Finally, we analyze the design of the rollout paradigms in \method, providing insights into why they outperform existing approaches and how they contribute to improved closed-loop simulation fidelity.

\subsection{Experimental Setup}
\paragraph{Simulation Environment}
All experiments are conducted in the CARLA simulator~\cite{carla}, following the RIFT framework~\cite{chen2025rift}, which facilitates AV-centric traffic simulation by assigning learned traffic policy to critical background vehicles (CBVs) that may interact with the autonomous vehicle (AV) along its designated route. RIFT adopts a two-stage learning pipeline: trajectory-level realism and route-level controllability are established through open-loop IL pretraining, while style-level diversity and covariate shift mitigation are subsequently addressed via RL fine-tuning.

To isolate the impact of rollout paradigms, the overall training pipeline, including the reward formulation, learning protocol, and evaluation procedure, is kept consistent with RIFT. The only modification lies in the forward simulation component, which is replaced by our proposed stepwise closed-loop paradigm. This controlled design enables an isolated and systematic examination of how rollout paradigms affect the realism and controllability of traffic flow.

\paragraph{Baselines}
To ensure fair comparison, we adopt the same baselines as RIFT, including RIFT itself as a strong reference. All methods are fine-tuned on the scoring head to isolate the effect of forward simulation. For prediction-based rollouts, the prediction head is also fine-tuned to ensure consistency between rollout and policy optimization.

\textbf{Pure RL/IL}: Models trained exclusively via RL or IL, including \textit{Pluto}~\cite{cheng2024pluto} and \textit{FREA}~\cite{chen2025frea}.

\textbf{RLFT/SFT}: Fine-tuning methods based on a pre-trained \textit{Pluto} model, including \textit{RIFT-Pluto}, \textit{PPO-Pluto}, \textit{REINFORCE-Pluto}, \textit{GRPO-Pluto}, and \textit{SFT-Pluto}.

\textbf{Hybrid}: Approaches that combine reinforcement and supervised fine-tuning, including \textit{RTR-Pluto} and \textit{RS-Pluto}.

All RL-based methods are trained under the normal style reward, following the standard configuration defined in RIFT.

\begin{table*}[t]
\renewcommand{\arraystretch}{1.3}
\centering
\caption{\small \textbf{Comparison in controllability and realism.} Metrics are evaluated under the PDM-Lite~\cite{Beibwenger2024pdmLite} AV setting across three random seeds, with the \textbf{best} and the \textcolor{DarkBlue}{\textbf{second-best}} results highlighted accordingly.}
\label{tab:main_results}
\vspace{-2mm}
\resizebox{1.0\textwidth}{!}{
\begin{threeparttable}
\tablestyle{2.0pt}{1.05}
\setlength{\tabcolsep}{1mm}{
\begin{tabular}{lcccccccccc}
    \toprule
     \multirow{3}{*}{\textbf{Method}} & \multirow{3}{*}{\textbf{Type}} & \multicolumn{3}{c}{\textbf{Kinematic}} & \multicolumn{4}{c}{\textbf{Interaction}}& \multicolumn{1}{c}{\textbf{Map}} & \multicolumn{1}{c}{\textbf{Comfort}}\\
     \cmidrule(r){3-5}
     \cmidrule(r){6-9}
     \cmidrule(r){10-10}
     \cmidrule(r){11-11}
     & & S-SW $\uparrow$ & S-WD $\downarrow$ & A-SW $\uparrow$ & CPK $\downarrow$ & RP $\uparrow$ & 2D-TTC $\uparrow$ & ACT $\uparrow$ & ORR $\downarrow$ & UC $\downarrow$\\
    \midrule
    Pluto~\cite{cheng2024pluto} & IL & {0.88} \pmsd {0.01} & {5.81} \pmsd {0.06} & {0.90} \pmsd {0.01} & {5.06} \pmsd {2.69} & {564.14} \pmsd {114.41} & {2.50} \pmsd {1.48} & {2.44} \pmsd {1.39} & {0.24} \pmsd {0.15} & \textcolor{DarkBlue}{\textbf{{56.45}}} \pmsd {4.14}\\
    FREA~\cite{chen2025frea} & RL & {0.93} \pmsd {0.01} & {5.10} \pmsd {0.14} & {0.93} \pmsd {0.01} & {30.42} \pmsd {5.28} & {292.81} \pmsd {68.54} & {2.71} \pmsd {1.40} & {2.67} \pmsd {1.41} & {9.01} \pmsd {2.09} & {72.40} \pmsd {1.72}\\
    \midrule
    SFT-Pluto & SFT & {0.88} \pmsd {0.02} & {6.01} \pmsd {0.19} & {0.87} \pmsd {0.02} & {6.33} \pmsd {2.23} & {780.48} \pmsd {41.05} & {2.20} \pmsd {1.64} & {2.12} \pmsd {1.51} & \textbf{{0.06}} \pmsd {0.07} & {68.14} \pmsd {4.91}\\
    RS-Pluto~\cite{peng2024improving} & SFT+RLFT & {0.93} \pmsd {0.00} & {5.40} \pmsd {0.15} & {0.92} \pmsd {0.01} & \textcolor{DarkBlue}{\textbf{{4.11}}} \pmsd {3.90} & {819.40} \pmsd {74.07} & {2.27} \pmsd {1.45} & {2.23} \pmsd {1.43} & {1.05} \pmsd {0.31} & {70.31} \pmsd {4.07}\\
    RTR-Pluto~\cite{zhang2023rtr} & SFT+RLFT & {0.85} \pmsd {0.00} & {6.24} \pmsd {0.16} & {0.81} \pmsd {0.03} & {6.98} \pmsd {2.59} & {481.60} \pmsd {70.19} & {2.55} \pmsd {1.60} & {2.47} \pmsd {1.51} & {0.08} \pmsd {0.09} & \textbf{{55.58}} \pmsd {4.76}\\
    PPO-Pluto & RLFT & {0.95} \pmsd {0.01} & {4.96} \pmsd {0.31} & {0.90} \pmsd {0.02} & {6.89} \pmsd {3.19} & {683.57} \pmsd {38.12} & {2.66} \pmsd {1.50} & {2.60} \pmsd {1.43} & \textcolor{DarkBlue}{\textbf{{0.07}}} \pmsd {0.13} & {58.29} \pmsd {2.70}\\
    REINFORCE-Pluto & RLFT & {0.92} \pmsd {0.01} & {5.63} \pmsd {0.19} & {0.90} \pmsd {0.02} & {6.98} \pmsd {0.86} & {813.70} \pmsd {24.76} & {2.39} \pmsd {1.64} & {2.30} \pmsd {1.55} & {1.37} \pmsd {1.13} & {68.10} \pmsd {1.22}\\
    GRPO-Pluto~\cite{shao2024grpo} & RLFT & {0.94} \pmsd {0.04} & {4.96} \pmsd {0.89} & \textbf{0.96} \pmsd {0.00} & {7.24} \pmsd {4.04} & {892.65} \pmsd {65.27} & {2.65} \pmsd {1.44} & {2.61} \pmsd {1.48} & {0.10} \pmsd {0.08} & {78.58} \pmsd {0.59}\\
    RIFT-Pluto~\cite{chen2025rift} & RLFT & \textbf{{0.97}} \pmsd {0.01} &  \textcolor{DarkBlue}{\textbf{{4.46}}} \pmsd {0.43} &  {0.93} \pmsd {0.01} & {6.83} \pmsd {2.62} & \textcolor{DarkBlue}{\textbf{{995.33}}} \pmsd {84.62} &  \textcolor{DarkBlue}{\textbf{{2.74}}} \pmsd {1.30} & \textcolor{DarkBlue}{\textbf{{2.71}}} \pmsd {1.32} & {0.36} \pmsd {0.20} & {76.90} \pmsd {2.82}\\
    \rowcolor{gray!25}ForSim-Pluto (ours) & RLFT & \textcolor{DarkBlue}{\textbf{{0.96}}} \pmsd {0.01} &  \textbf{{4.31}} \pmsd {0.72} &  \textcolor{DarkBlue}{\textbf{{0.94}}} \pmsd {0.01} & \textbf{{3.95}} \pmsd {1.57} & \textbf{{1005.15}} \pmsd {32.28} &  \textbf{{2.95}} \pmsd {1.40} & \textbf{{2.92}} \pmsd {1.46} & {0.34} \pmsd {0.31} & {57.12} \pmsd {3.03}\\
    \bottomrule
\end{tabular}
}
\end{threeparttable}
}
\vspace{-4mm}
\end{table*}

\subsection{Traffic Simulation Quality (Q1)}

\paragraph{Metrics}
Following the WOSAC evaluation framework~\cite{montali2023wosac}, we assess simulation quality using four categories of metrics: \textit{kinematic}, \textit{interaction}, \textit{map}, and \textit{comfort}.

\begin{table}[t]
% \begin{table}[t]
\caption{\small \textbf{Ablation over center rollout paradigms.} Entries with \textsuperscript{*} use Constant-Action for surrounding agents.}
\label{tab:center_ablation}
\vspace{-2mm}
\centering
\scriptsize
\tablestyle{2.0pt}{1.05}
\resizebox{0.49\textwidth}{!}{
\setlength{\tabcolsep}{1mm}{
\begin{tabular}{lcccc}
\toprule
\multirow{2}{*}{\textbf{Paradigms}} & \multicolumn{2}{c}{\textbf{Efficiency Metrics}} & \multicolumn{2}{c}{\textbf{Infraction Metrics}}\\
\cmidrule(r){2-3}
\cmidrule(r){4-5}
& BR $\downarrow$ & RP $\uparrow$ & ORR $\downarrow$ & CPK $\downarrow$\\
\midrule
Perfect Tracking\textsuperscript{*} &
\textbf{{0.00}} \pmsd {0.00} &
{993.63} \pmsd {48.01} &
\textcolor{DarkBlue}{\textbf{{0.26}}} \pmsd {0.22} &
\textcolor{DarkBlue}{\textbf{{4.00}}} \pmsd {0.83} \\
Trajectory Tracking\textsuperscript{*} &
{3.33} \pmsd {5.77} &
\textbf{{1087.56}} \pmsd {36.41} &
{0.67} \pmsd {0.57} &
{4.00} \pmsd {1.47} \\
\midrule
Max-Likelihood &
\textbf{{0.00}} \pmsd {0.00} &
{910.97} \pmsd {9.51} &
\textbf{{0.11}} \pmsd {0.05} &
{4.76} \pmsd {3.53} \\
Mode-Consistent &
{3.33} \pmsd {5.77} &
{940.94} \pmsd {73.70} &
{0.28} \pmsd {0.29} &
{4.55} \pmsd {1.34} \\
\rowcolor{gray!25} Trajectory-Aligned & 
\textbf{{0.00}} \pmsd {0.00} & 
\textcolor{DarkBlue}{\textbf{{1005.15}}} \pmsd {32.28} & 
{0.34} \pmsd {0.31} &
\textbf{{3.95}} \pmsd {1.57} \\
\bottomrule
\end{tabular}
}
}
% \vspace{-2mm}
% \end{table}
\vspace{-2mm}
% \begin{table}[t]
\caption{\small \textbf{Ablation over others rollout paradigms.}}
\label{tab:surr_ablation}
\vspace{-2mm}
\centering
\scriptsize
\tablestyle{2.0pt}{1.05}
\resizebox{0.49\textwidth}{!}{
\setlength{\tabcolsep}{1mm}{
\begin{tabular}{lcccc}
\toprule
\multirow{2}{*}{\textbf{Paradigms}} & \multicolumn{2}{c}{\textbf{Efficiency Metrics}} & \multicolumn{2}{c}{\textbf{Infraction Metrics}}\\
\cmidrule(r){2-3}
\cmidrule(r){4-5}
& BR $\downarrow$ & RP $\uparrow$ & ORR $\downarrow$ & CPK $\downarrow$\\
\midrule
Constant-Action &
\textbf{{0.00}} \pmsd {0.00} &
\textcolor{DarkBlue}{\textbf{{1128.77}}} \pmsd {96.24} &
\textcolor{DarkBlue}{\textbf{{0.91}}} \pmsd {0.30} &
{4.97} \pmsd {1.57} \\
Single-Prediction &
\textbf{{0.00}} \pmsd {0.00} &
\textbf{{1170.28}} \pmsd {29.17} &
{2.31} \pmsd {0.29} &
\textbf{{3.41}} \pmsd {1.44} \\
\rowcolor{gray!25} Stepwise Prediction & 
\textbf{{0.00}} \pmsd {0.00} & 
{1005.15} \pmsd {32.28} & 
\textbf{{0.34}} \pmsd {0.31} &
\textcolor{DarkBlue}{\textbf{{3.95}}} \pmsd {1.57} \\
\bottomrule
\end{tabular}
}
}
% \vspace{-4mm}
% \end{table}
\vspace{-5mm}
\end{table}

\textbf{Kinematic metrics} evaluate the distributional realism of agent motion, including the Shapiro–Wilk test on speed (S-SW) and acceleration (A-SW), and the Wasserstein distance on speed (S-WD).

\textbf{Interaction metrics} include Collision per Kilometer (CPK), Route Progress (RP), and safety-critical indicators: 2D-TTC~\cite{guo20232dttc} and ACT~\cite{venthuruthiyil2022act}.

\textbf{Map metric} measures spatial compliance via the Off-Road Rate (ORR), indicating time spent outside drivable areas.

\textbf{Comfort metric} captures driving smoothness via the Uncomfortable Rate (UC), defined as the proportion of time acceleration exceeds comfort thresholds, following RIFT~\cite{chen2025rift}.

\paragraph{Main Results and Analysis}
\method\ demonstrates marked improvements in safety-related metrics, notably reducing CPK, 2D-TTC, and ACT, as presented in \Cref{tab:main_results}. These results highlight the effectiveness of modeling closed-loop interactions, which enable agents to anticipate and respond to dynamic traffic conditions more effectively. Importantly, this enhanced safety is achieved without compromising efficiency or realism: kinematic characteristics and RP are preserved or even marginally improved relative to RIFT. At the same time, a reduced ORR suggests more spatially compliant behavior. Additionally, \method\ improves comfort substantially, benefiting from the use of the PID controller and the kinematic bicycle model that ensures physical plausibility.

\begin{figure*}[t]
  \centering
  \includegraphics[width=1.0\textwidth]{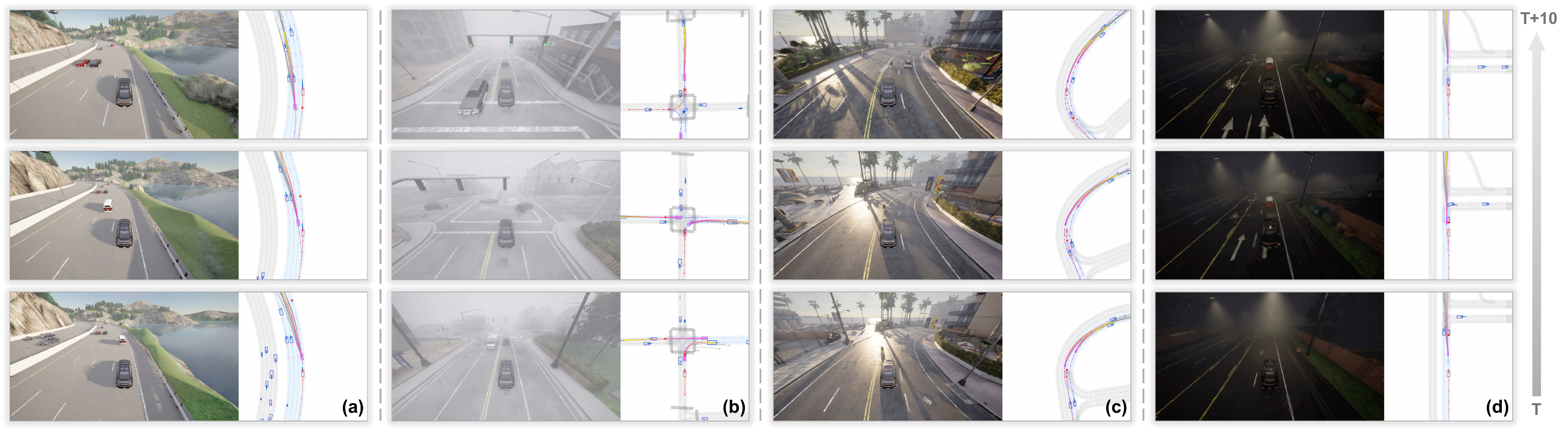}
  \vspace{-6mm}
  \caption{\textbf{Representative scenarios of \method.} The traffic agent (CBV) is marked in \highlighttransp{CBV-purple}{purple}, AV (PDM-Lite~\cite{Beibwenger2024pdmLite}) is in \highlighttransp{AV-red}{red}, and other agents are in \highlighttransp{BV-blue}{blue}.} 
  \label{fig:real_vis}
  \vspace{-2mm}
\end{figure*}

\subsection{Rollout Paradigm Analysis (Q2 \& Q3)}

\paragraph{Ablation over Rollout Paradigms}
We conduct ablation studies over both traffic agent and other agent rollout paradigms, as shown in \Cref{tab:center_ablation,tab:surr_ablation}. For the traffic agent (\Cref{tab:center_ablation}), we compare three closed-loop variants: Max-Likelihood Rollout, Mode-Consistent Rollout, and Trajectory-Aligned Rollout, while fixing other agents to the Stepwise Prediction Rollout. In contrast, Perfect Tracking and Trajectory Tracking bypass the traffic policy during forward simulation and are unaffected by other agent transitions. We therefore retain the standard Constant-Action Rollout for other agents, consistent with prior work~\cite{dauner2023tuplan_garage}.

Among traffic agent paradigms, the Trajectory-Aligned Rollout offers the most favorable trade-off between safety and efficiency, achieving the lowest BR and CPK with competitive RP and ORR. By selecting the spatiotemporally aligned trajectory that best matches the initial reference, this strategy enforces physical modality consistency. When integrated with stepwise rollout, it captures interaction-aware transitions that more faithfully emulate real-world multi-agent interactions.

For other agents (\Cref{tab:surr_ablation}), we fix the traffic agent to the Trajectory-Aligned Rollout and evaluate three rollout paradigms: Constant-Action Rollout, Single-Prediction Rollout, and Stepwise Prediction Rollout. Constant-Action Rollout, as an open-loop strategy, neglects evolving traffic dynamics, resulting in poor responsiveness and higher CPK. Single-Prediction Rollout improves short-horizon accuracy but lacks long-horizon supervision, leading to error accumulation and frequent off-road deviations, reflected in higher ORR. In contrast, Stepwise Prediction Rollout leverages short-horizon accuracy in a closed-loop manner, continuously adapting the agent's behavior to dynamic contexts, thereby achieving competitive ORR and CPK while maintaining high RP and eliminating blocking. These findings underscore the effectiveness of closed-loop modeling in capturing responsive and interactive multi-agent behavior.

Together, these results show that combining the Trajectory-Aligned Rollout with Stepwise Prediction Rollout is the most effective forward simulation configuration, substantially improving safety while preserving efficiency.

\paragraph{Representative Scenarios Analysis}
\Cref{fig:real_vis} presents representative scenarios that highlight the effectiveness of \method\ under the proposed forward simulation paradigm. By modeling closed-loop virtual interactions, the traffic policy learns to select context-aware, risk-sensitive behaviors.

\begin{figure}[t]
  \centering
  \includegraphics[width=0.48\textwidth]{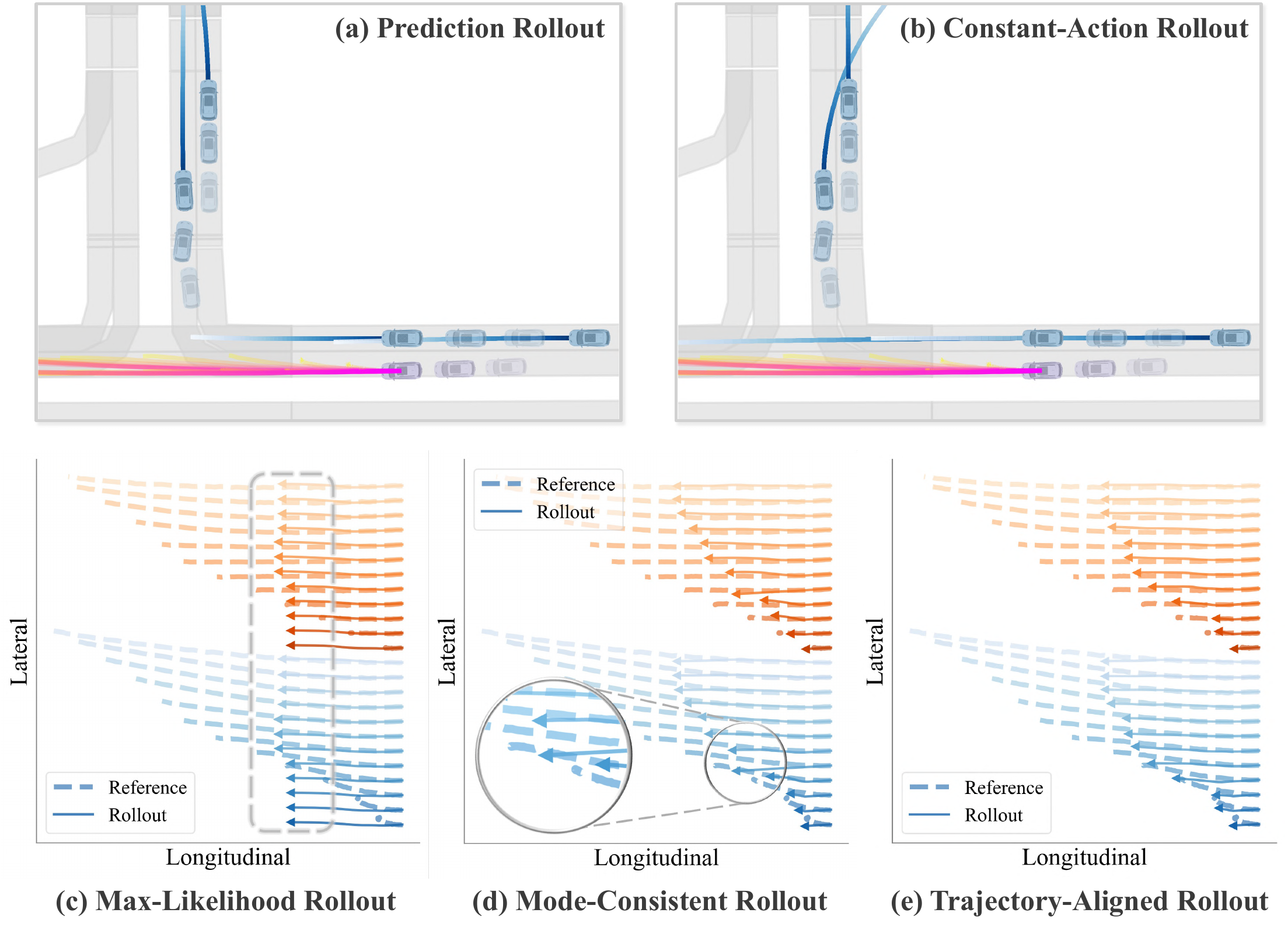}
  \vspace{-5mm}
  \caption{\textbf{Comparison of rollout paradigms.} For other agents, (a)–(b): Prediction Rollout improves long-horizon stability over Constant-Action Rollout. For the traffic agent, (c)–(e): Trajectory-Aligned Rollout preserves multimodal coherence, while Max-Likelihood Rollout and Mode-Consistent Rollout suffer from mode collapse and temporal inconsistency.}
  \label{fig:virtual_vis}
  \vspace{-5mm}
\end{figure}

In \Cref{fig:real_vis}(a), the agent performs a high-efficiency lane-change strategy by shifting from the outer to the inner lane during a low-curvature turn, highlighting its ability to anticipate and optimize future motion. \Cref{fig:real_vis}(b) illustrates that, despite multiple off-road candidates near an intersection, the agent selects a reference-aligned trajectory that adheres to spatial constraints. \Cref{fig:real_vis}(c) presents two cyclists recovering from off-center positions during a turn, demonstrating robustness to covariate shift and the capacity for corrective behavior. \Cref{fig:real_vis}(d) shows an overtaking maneuver initiated in response to a lead vehicle's deceleration, where the agent avoids potential collision while preserving driving efficiency.

These cases collectively demonstrate that modeling closed-loop interactions during forward simulation enables the traffic policy to internalize risk-aware, efficient, and realistic behaviors across diverse scenarios.

\paragraph{Qualitative Analysis of Rollout Paradigms}

\Cref{fig:virtual_vis} provides qualitative comparisons of rollout paradigms. \Cref{fig:virtual_vis}(a)–(b) contrast two rollout paradigms for other agents: Prediction Rollout and Constant-Action Rollout. The former yields more stable and accurate long-horizon rollouts, as its trained predictor generates physically plausible futures and is less sensitive to transient control noise. In contrast, the Constant-Action Rollout propagates only the initial action throughout the horizon, resulting in severe error accumulation and reduced realism.

\Cref{fig:virtual_vis}(c)–(e) visualize traffic agent rollouts under three paradigms. The Max-Likelihood Rollout, as discussed in \Cref{subsec:center_rollout}, collapses diverse intention modes by repeatedly selecting the highest-probability trajectory, limiting diversity and reducing intra-group advantage separation. The Mode-Consistent Rollout enforces explicit modal consistency by fixing the selected mode, but suffers from temporal inconsistency—trajectories from the same mode across different virtual states may not be physically aligned, leading to modality drift. In contrast, the Trajectory-Aligned Rollout enforces physical consistency by selecting the candidate that best matches the reference trajectory in spatiotemporal distance. This approach preserves multimodal diversity while enhancing physical plausibility and intra-modality consistency.

\section{Conclusions}
\label{sec:conclusion}

In this work, we proposed \method, a stepwise closed-loop forward-simulation paradigm that addresses key limitations in existing traffic simulation frameworks, particularly their non-reactivity and lack of interactive fidelity in prior rollout procedures. \method\ employs the Trajectory-Aligned Rollout for traffic agents at each virtual timestep and propagates them with physically grounded dynamics, thereby preserving multimodal diversity, intra-modality consistency, and physical plausibility. Meanwhile, other agents are updated via Stepwise Prediction Rollout, enabling responsive, interaction-aware evolution under closed-loop conditions. Integrated into the RIFT framework, \method\ facilitates interaction-aware simulation and fine-grained policy optimization via group-relative objectives. Experiments demonstrate that \method\ improves safety-related metrics without compromising efficiency, realism, or comfort, validating the importance of modeling closed-loop multimodal interactions in virtual simulation. These results establish \method\ as a general and effective paradigm for enhancing the fidelity and reliability of traffic simulation.
% \section*{ACKNOWLEDGMENT}

\bibliographystyle{IEEEtran}
\bibliography{IEEEabrv,ref}

\end{document}